\documentclass[runningheads]{llncs}
\usepackage[T1]{fontenc}
\usepackage{cite}
\usepackage{amsmath,amssymb,amsfonts}
\usepackage{wrapfig}
\usepackage{algorithmic}
\usepackage{graphicx}
\usepackage{textcomp}
\usepackage{xcolor}
\usepackage{xurl}
\usepackage{hyperref}
\usepackage{appendix}
\usepackage{lipsum}
\usepackage{multirow}
\usepackage{multicol}
\usepackage{tabularx}
\usepackage{caption}
\usepackage{subcaption}
\usepackage[misc]{ifsym}

\usepackage{xcolor,booktabs}
\definecolor{light-gray}{gray}{0.95}

\makeatletter
\newcommand{\printfnsymbol}[1]{%
  \textsuperscript{\@fnsymbol{#1}}%
}
\makeatother

\begin{document}
\title{A Proposed Large Language Model-Based Smart Search for Archive System}
%
%
\author{Ha Dung Nguyen \inst{1,2,3,}\thanks{These authors contributed equally to this work}\textsuperscript{(\Letter)} \and
Thi-Hoang Anh Nguyen\inst{1,2,3,}\printfnsymbol{1} \and
Thanh Binh Nguyen\inst{1,2,3}}
\authorrunning{Dung and Anh et al.}
%
\institute{
Vietnam National University, HCM \and
University of Information Technology, Ho Chi Minh, Vietnam \and
\email{dungngh@uit.edu.vn, anhnth@uit.edu.vn, binhnt@uit.edu.vn}}

\setcounter{secnumdepth}{5}

\maketitle
\begin{abstract}
This study presents a novel framework for smart search in digital archival systems, leveraging the capabilities of Large Language Models (LLMs) to enhance information retrieval. By employing a Retrieval-Augmented Generation (RAG) approach, the framework enables the processing of natural language queries and transforming non-textual data into meaningful textual representations. The system integrates advanced metadata generation techniques, a hybrid retrieval mechanism, a router query engine, and robust response synthesis, the results proved search precision and relevance. We present the architecture and implementation of the system and evaluate its performance in four experiments concerning LLM efficiency, hybrid retrieval optimizations, multilingual query handling, and the impacts of individual components. Obtained results show significant improvements over conventional approaches and have demonstrated the potential of AI-powered systems to transform modern archival practices.

\end{abstract}
\section{Introduction}

In the information age, the proliferation of digital data has transformed the nature of archival systems from traditional, static repositories of physical documents to dynamic, interactive environments capable of storing, processing, and retrieving vast amounts of information. Digital archives encompass a variety of media types, including text, images, audio, and video, necessitating advanced methods for effective data management and retrieval. As users increasingly demand seamless access to diverse forms of information, traditional keyword-based search methodologies often fall short \cite{10.1371/journal.pone.0293034, FERNANDEZ2011434}, leading to irrelevant results and an inability to grasp the nuances of user queries. To address these challenges, this study proposes a framework for smart search in archival systems that leverages the strengths of LLMs.

To facilitate semantic understanding and enhance cross-modal search capabilities, we designed our framework based on a RAG architecture. This architecture integrates the retriever's similarity-based file searching with the language comprehension capabilities of an LLM. Our system leverages existing LLM architectures, specifically utilizing llamaindex \footnote{\url{https://www.llamaindex.ai/}} for its robust RAG functionalities. This approach allows for the generation of responses directly from user data, bypassing the need for extensive fine-tuning of the LLM, which is both time-consuming and resource-intensive.

\begin{figure}[ht]
    \centering
    \caption{Illustration of User Input and Corresponding Desired Output in Our Proposed System}
    \label{fig:example}
    \includegraphics[width=\linewidth]{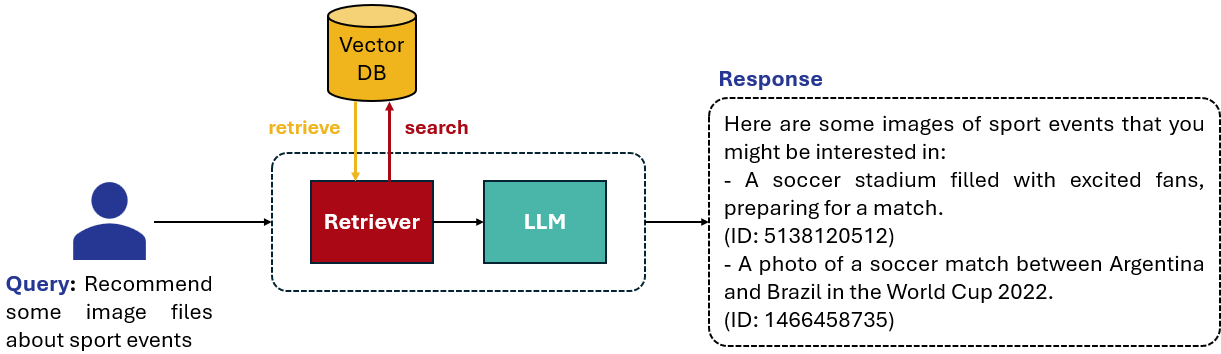}
\end{figure}

Figure \ref{fig:example} illustrates the processing of a typical user query within our system and the structured output generated. Unlike conventional databases that deliver fixed outputs, our system is engineered to produce responses that are both human-like in their readability and well-structured, always including file IDs for easy reference. This flexibility enables customization of the metadata returned, whether as file paths or direct URLs to the files.

The significance of this research lies in its potential to revolutionize the way digital archives function, making them more responsive to user needs and better equipped to handle the complexities of multimodal data. By harnessing LLMs, the proposed framework aims to redefine archival search technologies, ensuring that users can efficiently access and utilize the wealth of information contained within digital archives in the digital age.

\vspace*{-0.5cm}
\vspace*{-1cm}

\section{Backgrounds and Preliminaries}

\subsection{Archive System}
Digital archives have gained prominence in many fields, reflecting the growing reliance on databases in the information age \cite{artexte19207}. Traditional archives, rooted in print and paper-based systems, differ from digital archives, which encode texts, sounds, and images as data. This shift has transformed archives from static repositories of documents into dynamic systems that store, process, and circulate information \cite{lirias1868258}. Digital data derives value from its potential for future use, converting raw data into meaningful information through various processes \cite{Manovich.2001}. Unlike traditional archives, which focus on storage, digital archives emphasize continuous updates and user-centric information.

A digital archive system consists of key components for storing, organizing, and retrieving records. Storage relies on secure, scalable digital solutions like cloud storage, with redundancy and backups for data preservation. Organization is handled through metadata tagging, hierarchical structures, and indexing to ensure efficient management and retrieval. Retrieval systems are user-friendly, combining traditional keyword searches with advanced techniques like natural language search to enhance access to records.

\subsection{Search Features in Archive System}

\subsubsection{Traditional Search}

Traditional systems of archiving depend on keyword-based or metadata-driven methods of searching. The user enters specific terms or selects metadata attributes, like title, author, date, or file type, and a set of matching records is returned. Most of these represent simple textual-matching algorithms which scan the database of the archive for records with specified keywords or metadata values. They'll be fairly effective within well-managed archives where metadata is consistently applied. In contrast, some of the limits to traditional search are irrelevant results, or missing the most important; understanding what queries and content mean; to handle complex queries.

\subsubsection{LLM-based Search}

LLM-based search enhances retrieval by leveraging Large Language Models to interpret context and query semantics, surpassing traditional methods that rely on physical metadata or user-generated descriptions. This approach offers semantic understanding, enabling better interpretation of user intent; facilitates cross-modal search by unifying descriptions across media types; and supports natural language interaction, making searches more intuitive and effective across formats.

\subsection{Metadata in Archive System}

Metadata is structured information that describes data in context, providing better usability of data through a bridge between it and the users. In an archive system, it plays a crucial role in discovery, management, and understanding. For its discovery, it aids in the locating of relevant assets by indexing and categorizing data attributes. In terms of management, metadata functions to systematically organize, preserve, and control digital assets. It finally enhances understanding by providing context, hence enabling the user to interpret data more easily.

Archival metadata can be divided into three categories, which in many cases adhere to established guidelines such as the Dublin Core or the Europeana Data Model \cite{meta01}. These categories include physical metadata, custom metadata, and AI-generated metadata. \textbf{Physical metadata} would contain technical specifications such as format, size, and creation date which allows sufficient system integration and protection of data integrity. \textbf{Custom metadata}, often user-generated, is tailored to meet the specific needs of an archive, enhancing categorization and retrieval based on the unique requirements of the archive. \textbf{AI metadata} is the opposite of custom metadata wherein the AI does the work and generates the content, object recognition, transcription of speech-to-text, sentiment appraisal etc., thus booting advanced search and utilization of data.

\section{A Proposed Framework for Smart Search in Archive System}

\subsection{Knowledge Base Creation} \label{sec:data_preparation}

Our system is similar to a RAG framework: the user provides a natural language query, and then the system retrieves information from its knowledge base and generates an answer. A knowledge base refers to structured information that a system uses while understanding queries and returning appropriate responses. Thus, the first critical step is building a robust knowledge base to support the system's operations.

\subsubsection{Data Collection and Preparation}

Various sources, including databases, external archives, and web scraping, are employed to gather relevant data tailored to user needs. The knowledge base comprises diverse data types, including text, images, audio, and video; however, all non-textual data is represented as uniform text to streamline processing. AI-generated content can effectively enrich this representation, with AI agents generating short descriptions for images, converting audio content into text, and producing descriptive summaries for videos by analyzing their visual and audio components. This approach enables the capture and storage of multimodal information efficiently, minimizing the storage requirements associated with actual non-textual files. Additionally, metadata, including physical attributes, domain-specific annotations, and AI-generated insights, plays a crucial role in organizing and enhancing the accessibility of the collected data.

\subsubsection{Embedding Creation and Indexing}

This process involves converting data into vector representations that capture the semantic meaning of the content. Once transformed, these embeddings are securely stored in a vector database, allowing for efficient similarity-based retrieval by comparing query vectors with the stored data vectors. To handle diverse data types and large-scale datasets, the system often divides data into smaller, semantically meaningful chunks. This structured indexing facilitates accurate retrieval, ensuring relevant responses to user queries.

In this study, we utilize the BGE-M3 embedding model \cite{bge_m3} to generate high-quality vector representations. For indexing and retrieval, we employ Pinecone, a scalable vector database optimized for real-time updates and high-performance similarity searches. Other embedding models and vector databases could be used based on the specific user case and preference of the user.

\subsection{The Architecture of the Proposed Smart Search System.}

\begin{figure}[ht]
    \centering
    \caption{The Architecture of the Proposed Smart Search System.}
    \label{fig:framework}
    \includegraphics[width=\linewidth]{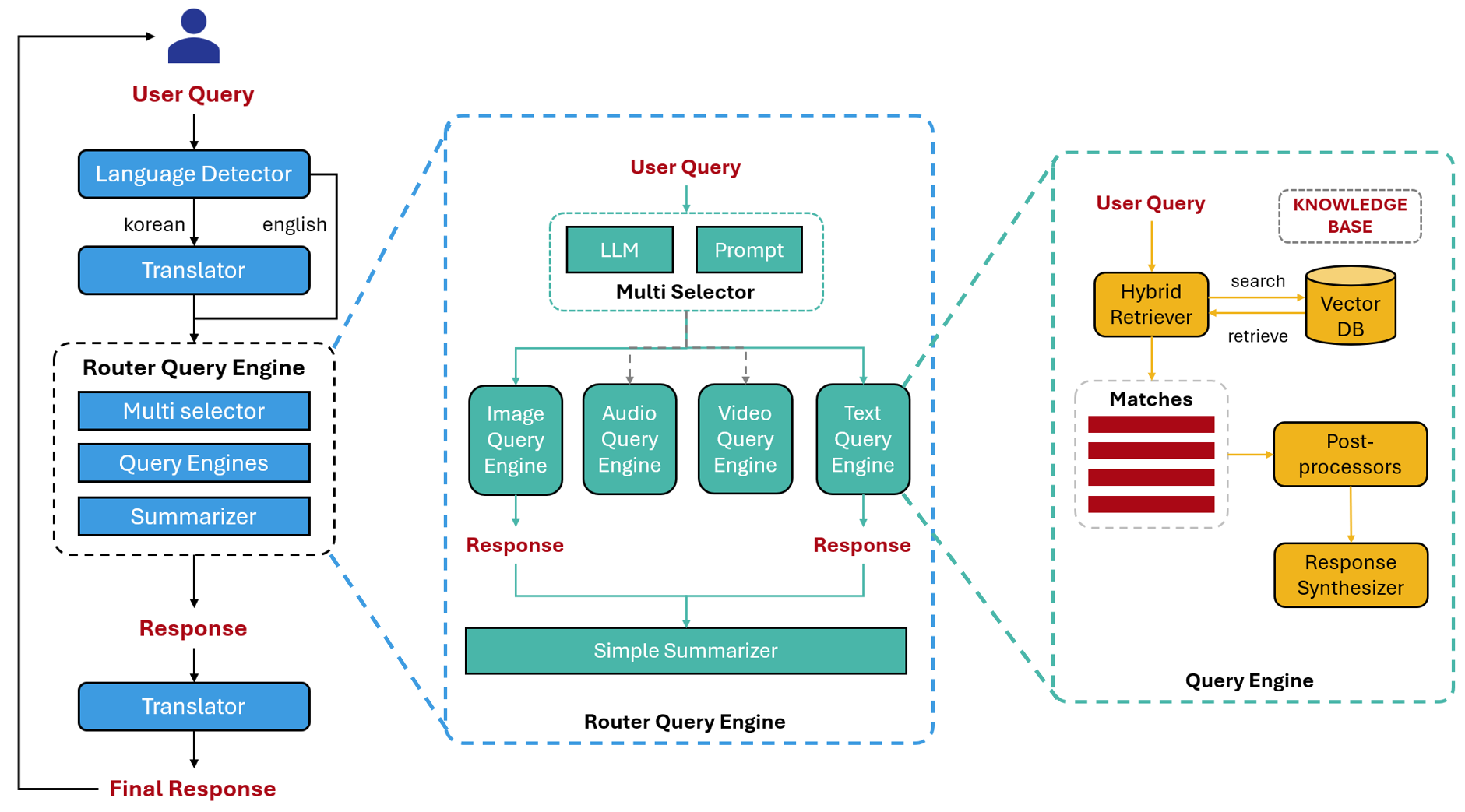}
\end{figure}

The framework presented (Figure \ref{fig:framework}) draws inspiration from various Retrieval-Augmented Generation (RAG) studies \cite{gao2024retrievalaugmentedgenerationlargelanguage, fan2024surveyragmeetingllms, ram2023incontextretrievalaugmentedlanguagemodels}, meticulously crafting an architecture tailored for efficient multimedia data handling. It integrates robust data retrieval and response generation components. Novel solutions within this framework include hybrid retrieval techniques and query routing capabilities, ensuring precise multimedia data retrieval and the delivery of coherent, user-friendly responses.

\subsubsection{Translator (Query and Response)}

This component detects the language of user queries and translates non-English queries into English for processing. After generating a response, it is translated back into the original query language, ensuring users receive replies in their preferred language. We used the Detect Language API \footnote{\url{https://detectlanguage.com/}} for language detection and the GoogleTranslator module from the deep-translator library \footnote{\url{https://pypi.org/project/deep-translator/}} for translations.



\subsubsection{Router Query Engine}

The Router Query Engine is designed to handle different types of multimedia data by routing queries to the appropriate specialized query engine based on the input type. It integrates tools tailored for images, audio, video, and documents, each managed by its query engine. The engine uses an intelligent selector to decide which tool to apply, ensuring accurate and relevant responses. A summarization component (simple summarizer) further refines the output, providing users with a concise, informative result regardless of the media type. This system optimizes query handling across diverse data sources efficiently.

\subsubsection{Hybrid Retriever}

\begin{figure}[ht]
    \centering
    \caption{Hybrid Retriever.}
    \label{fig:hybrid}
    \includegraphics[width=\linewidth]{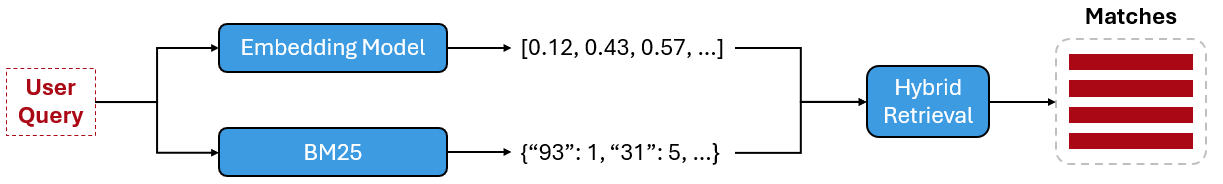}
\end{figure}

A hybrid retriever, as illustrated in Figure \ref{fig:hybrid}, combines the strengths of traditional keyword-based retrieval methods, in our case BM25, with modern vector-based approaches. This model has been proven to enhance search accuracy by leveraging both keyword specificity and the semantic richness of vector embeddings \cite{sawarkar2024blendedragimprovingrag, mandikal2024sparsemeetsdensehybrid}.

A critical aspect of this integration is the alpha parameter ($\alpha$), which plays a pivotal role in weighting the contributions of each method to the final retrieval score. As shown in Equation \ref{eq:hybrid}, alpha controls the balance between the BM25 score, which emphasizes exact keyword matches, and the semantic score derived from the vector embeddings that capture contextual meanings. When alpha is set closer to 0, the hybrid retriever prioritizes BM25, emphasizing precision in keyword matches. Conversely, when alpha approaches 1, the retriever leans towards vector-based retrieval, enhancing semantic generalization. By tuning alpha, the system can be adjusted to handle varying levels of query complexity, optimizing precision, recall, and overall retrieval effectiveness across diverse datasets.

\begin{equation}
    \text{Hybrid Score} = \alpha \cdot \text{Embedding Model Score} + (1 - \alpha) \cdot \text{BM25 Score}
    \label{eq:hybrid}
\end{equation}

\subsubsection{Post-Processors}

The post-processors component enhances retrieved results by employing two key techniques: reranking and long context reordering. Reranker \footnote{\url{https://huggingface.co/cross-encoder/ms-marco-MiniLM-L-2-v2}} re-evaluates, scores, and filters retrieved nodes to prioritize the most relevant information \cite{Gao_2022,tao-etal-2023-core}, while long context reorder optimizes the handling of lengthy contexts by rearranging nodes to focus the language model's attention on key information \cite{liu2023lostmiddlelanguagemodels}. This ensures that the response synthesizer receives refined and relevant results, leading to more accurate and informative responses.

\subsubsection{Response Synthesizer}

\begin{figure}[ht]
    \centering
    \caption{Response Synthesizer.}
    \label{fig:synthesizer}
    \includegraphics[width=\linewidth]{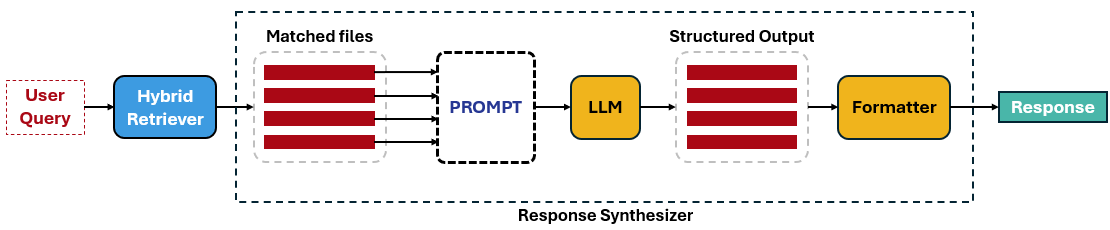}
\end{figure}

The response synthesizer effectively integrates results from the hybrid retriever and utilizes an LLM to generate comprehensive and structured responses. By applying a consistent prompt to each retrieved chunk, the synthesizer ensures a cohesive and coherent output. Furthermore, its ability to dynamically adapt the response format to match the specific data type guarantees that the information provided is not only accurate but also presented in a manner that is easily understandable and applicable. This adaptability is crucial for ensuring that the generated responses are not only informative but also contextually relevant and valuable to the user. Figure \ref{fig:synthesizer} visually depicts the operation of our synthesizer.

\section{Experiments Settings}

Our experiments focus on evaluating the proposed system’s ability to accurately include relevant file IDs in the LLM-generated responses, rather than simply assessing raw retrieval performance. This simulates real-world scenarios where users expect clear references to specific files within the response content.

\subsection{Experimental Dataset}

To comprehensively assess the system across varied scenarios, three distinct types of queries were crafted to emulate file retrieval tasks from the archive, each designed to retrieve files based on their type(s) and specified topic:

\begin{itemize}
    \item \textbf{Type 1}: "Recommend some \{filetype\} files about \{topic\}" (Query a specific file type on a topic)
    \item \textbf{Type 2}: "Retrieve some \{filetype1\} or \{filetype2\} files about \{topic\}" (Query two specific file types on a topic)
    \item \textbf{Type 3}: "Give me some files about \{topic\}" (Query files with any type on a topic)
\end{itemize}

Utilizing four file types (image, audio, video, text document), and ten topics, such as wildlife, landscapes, celebrities, and political events, we generated a total of 110 queries (type 1: 40, type 2: 60, and type 3: 10 queries) to encompass a broad range of retrieval scenarios.

We utilized a variety of LLMs to create diverse data files across all file types and topics. These files were subsequently indexed into Pinecone, organized into four separate indices corresponding to each file type. The experimental dataset is made available on our Github repository \footnote{Github: \url{https://github.com/AnhHoang0529/archive_smart_search}}.

\subsection{Experiment Design}

Four key experiments were designed to assess various aspects of the RAG system's performance:

\begin{enumerate}
    \item \textbf{Comparision of LLM performance}: We compared the performance of different multilingual LLMs, all of which have 7 billion parameters: Mistral 7B \cite{jiang2023mistral7b}, Synatra 7B \footnote{\url{https://huggingface.co/maywell/Synatra-7B-Instruct-v0.2}}, and Llama 2 7B \cite{touvron2023llama2openfoundation}. The goal was to evaluate the quality of responses generated by each LLM and determine which model most effectively includes relevant file IDs in its outputs.
    \item \textbf{Hybrid Retriever Adjustment}: We evaluated a hybrid retrieval mechanism combining BM25 and Vector Retriever, using an alpha parameter to control their contribution. We tested alpha values from 0.0 to 1.0 to examine performance trade-offs.
    \item \textbf{Multilingual Query Testing}: This experiment evaluated the performance of Mistral, Synatra, and Llama 2 in handling English and Korean queries, assessing the system's ability to manage multilingual queries.
    \item \textbf{Component Reduction}: We conducted an ablation study to assess the impact of removing key components from the RAG system. By excluding the translator, query router, and post-processor components from the proposed platform, we evaluated the role of translation, query routing, and refinement in overall performance.
\end{enumerate}

\subsection{Evaluation Metrics}

To evaluate the performance of our proposed system, we utilize four key metrics: \textbf{(1) Precision} - Proportion of retrieved files that are relevant, \textbf{(2) Recall} - Proportion of relevant files retrieved, \textbf{(3) F1-score} - Harmonic mean of precision and recall, and \textbf{(4) Hit Rate} - Percentage of queries with at least one relevant file retrieved.

\begin{figure}[ht]
    \centering
    \caption{Illustration of retrieved files from the proposed system: In this example, although the retriever extracted four files, only two files are mentioned in the LLM's response. Consequently, the retrieved files are identified as 5138120512 and 1466458735.}
    \label{fig:retrieved_files}
    \includegraphics[width=\linewidth]{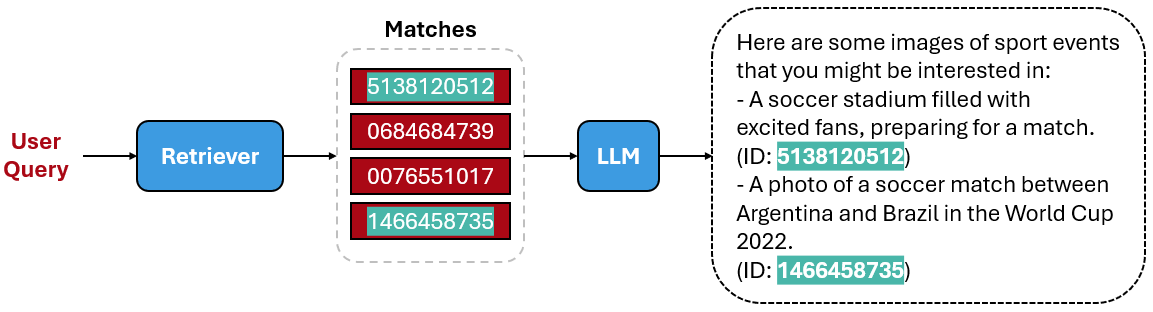}
\end{figure}

It is important to note that in our research, the retrieved files are those referenced in the LLM's response text, rather than files directly retrieved from the underlying retrieval system, as illustrated in Figure \ref{fig:retrieved_files}.

\section{Experiment Results and Analysis}

\subsection{Experiment 1: Comparision of LLM performance}

As shown in Table \ref{tab:exp1-metrics}, Mistral significantly outperformed Synatra and Llama 2 across key metrics. It achieved a high average precision of 80.56\% and a strong balance between precision and recall. Synatra excelled in complex multi-filetype queries but struggled with recall in general queries. Llama 2 exhibited the weakest performance, with a low average precision and recall.

\begin{figure}[ht]
    \centering
    \includegraphics[width=0.8\linewidth]{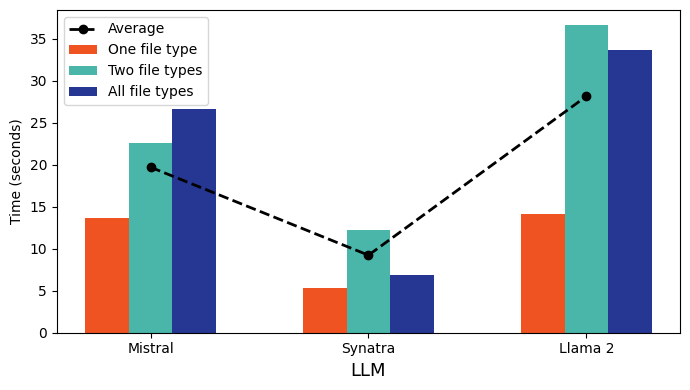}
    \caption{Time Performance of LLMs by Query Type.}
    \label{fig:exp1}
\end{figure}

Figure \ref{fig:exp1} shows that Synatra is the fastest LLM, significantly outperforming Mistral and Llama 2. Mistral has a moderate execution time, while Llama 2 is the slowest, particularly struggling with broader queries. These results suggest that Synatra is more efficient in generating responses, while Mistral offers a balance between execution time and performance, and Llama 2 may pose latency challenges.

\begin{table}[]
\centering
\caption{Performance Metrics Comparison of Multilingual LLMs for File Retrieval}
\label{tab:exp1-metrics}
\begin{tabularx}{\linewidth}{>{\centering\arraybackslash}X>{\centering\arraybackslash}X>{\centering\arraybackslash}X>{\centering\arraybackslash}X>{\centering\arraybackslash}X>{\centering\arraybackslash}X}
\toprule
\multirow{2}{*}{\textbf{LLM}}     & \multirow{2}{*}{\textbf{Query type}} & \multicolumn{4}{c}{\textbf{Metrics (\%)}}                                         \\
                                  &                                      & \textbf{Precision} & \textbf{Recall} & \textbf{F1-score} & \textbf{Hit rate} \\
\midrule
\multirow{4}{*}{\textbf{Mistral}} & One filetype                         & 80.25              & 61.99           & 68.70             & 95.00             \\
                                  & Two filetypes                        & 79.82              & 59.65           & 67.12             & 91.67             \\
                                  & All filetypes                        & 86.25              & 37.23           & 50.83             & 100.0             \\
                                  & \textbf{Average}                     & \textbf{80.56}     & \textbf{58.46}  & \textbf{66.21}    & \textbf{93.64}    \\
\midrule
\multirow{4}{*}{\textbf{Synatra}} & One filetype                         & 34.54              & 42.38           & 37.38             & 50.00             \\
                                  & Two filetypes                        & 66.37              & 43.30           & 51.68             & 95.00              \\
                                  & All filetypes                        & 84.44              & 18.99           & 30.54             & 100.0              \\
                                  & \textbf{Average}                     & \textbf{56.44}     & \textbf{40.75}  & \textbf{44.6}     & \textbf{79.09}    \\
\midrule
\multirow{4}{*}{\textbf{Llama 2}} & One filetype                         & 37.87              & 44.40           & 38.57             & 67.50             \\
                                  & Two filetypes                        & 61.64              & 34.38           & 42.34             & 93.33             \\
                                  & All filetypes                        & 43.99              & 13.98           & 20.32             & 60.00              \\
                                  & \textbf{Average}                     & \textbf{51.39}     & \textbf{36.17}  & \textbf{38.97}    & \textbf{80.91}   \\
\bottomrule
\end{tabularx}
\end{table}
\vspace*{-0.5cm}

\subsection{Experiment 2: Hybrid Retriever Adjustment}

\begin{figure}[h]
     \centering
     \caption{Performance Comparison of Hybrid Retriever with multiple alpha values.}
     \label{fig:exp2}
     \begin{subfigure}[b]{\textwidth}
         \centering
         \includegraphics[width=\textwidth]{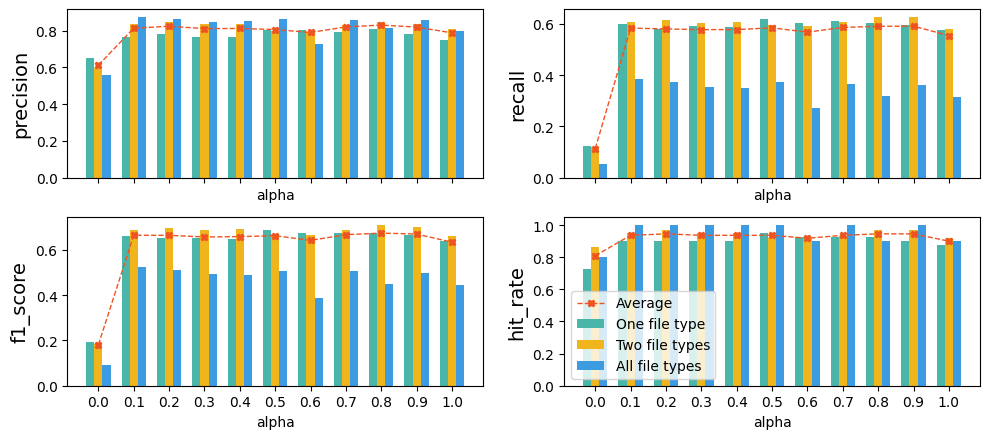}
         \caption{Performance Metrics.}
         \label{fig:exp2_metrics}
     \end{subfigure}
     \hfill
     \begin{subfigure}[b]{\textwidth}
         \centering
         \includegraphics[width=\textwidth]{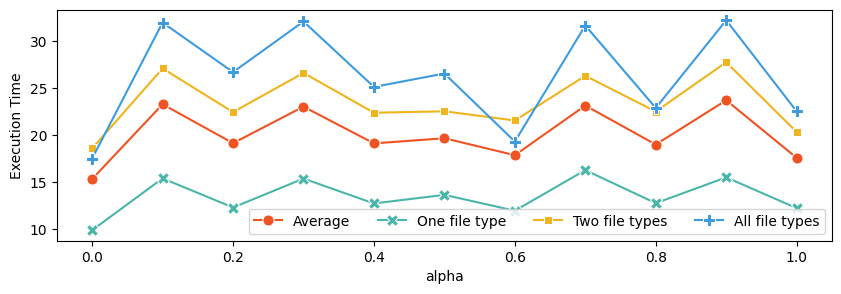}
         \caption{Execution Time Comparison.}
         \label{fig:exp2_time}
     \end{subfigure}
\end{figure}

As shown in Figure \ref{fig:exp2_metrics}, precision remained consistently high across the tested alpha values, reaching a peak of 83\% with an alpha of 0.8. However, precision slightly decreased at alpha 1.0. The recall followed a similar trend, peaking around alpha 0.9, while the F1 score also achieved its maximum value at alpha 0.8, reaching 67.34\%. The hit rate remained stable between 93\% and 94\%, showing minimal variation with different alpha values. Regarding execution time, depicted in Figure \ref{fig:exp2_time}, a clear pattern emerged: even alpha values tended to have lower execution times than odd alpha values.

In general, alpha values of 0.5 or 0.8 represent an optimal balance between retrieval accuracy and execution time. These settings effectively leverage the precision of BM25 and the contextual understanding of vector search, optimizing overall system performance.

\subsection{Experiment 3: Multilingual Query Testing}

As illustrated in Table \ref{tab:exp3}, the performance difference between English and Korean queries is notable across all models. For Mistral, precision and F1 scores are higher for English queries (80.56\% and 66.21\%) compared to Korean (71.61\% and 59.67\%). Similarly, Synatra and Llama 2 exhibit better performance on English queries, with both models showing lower recall and F1 scores for Korean. This highlights a consistent drop in performance when processing Korean queries across all models. The proposed system handles multilingual queries but shows reduced performance for Korean, indicating a need for improvement in non-English language processing.


\begin{table}[ht]
\centering
\caption{Performance Metrics of LLMs by Query's Language}
\label{tab:exp3}
\begin{tabularx}{\linewidth}{>{\centering\arraybackslash}X>{\centering\arraybackslash}X>{\centering\arraybackslash}X>{\centering\arraybackslash}X>{\centering\arraybackslash}X>{\centering\arraybackslash}X}
\toprule
\multirow{2}{*}{\textbf{LLM}}       & \multirow{2}{*}{\textbf{Language}} & \multicolumn{4}{c}{\textbf{Metrics   (\%)}}                                  \\
                                    &                                    & \textbf{Precision} & \textbf{Recall} & \textbf{F1 score} & \textbf{Hit Rate} \\
\midrule
\multirow{2}{*}{\textbf{Mistral}}   & English                                 & 80.56              & 58.46           & 66.21             & 93.64             \\
                                    & Korean                                 & 71.61              & 54.69           & 59.67             & 90.91             \\
\midrule
\multirow{2}{*}{\textbf{Synatra}}   & English                                 & 56.44              & 40.75           & 44.56             & 79.09             \\
                                    & Korean                                 & 50.66              & 38.59           & 40.84             & 78.18             \\
\midrule
\multirow{2}{*}{\textbf{Llama   2}} & English                                 & 51.39              & 36.17           & 38.97             & 80.91             \\
                                    & Korean                                 & 41.35              & 31.84           & 33.46             & 70.91             \\
\bottomrule
\end{tabularx}
\end{table}

\subsection{Experiment 4: Ablation Study}



\begin{figure}[ht]
    \centering
    \includegraphics[width=0.8\linewidth]{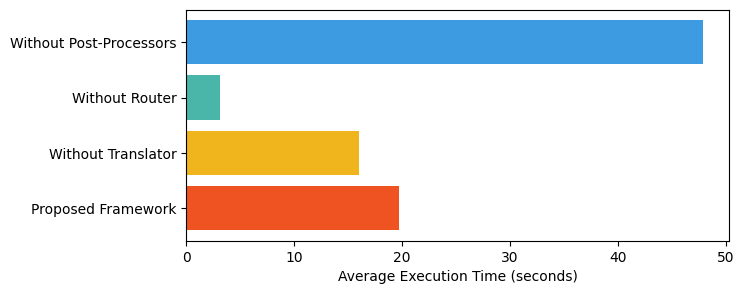}
    \caption{Execution Time Comparison for Ablation Study on the Proposed Framework.}
    \label{fig:exp4}
\end{figure}

Table \ref{tab:exp4} highlights the importance of each component in our system. Removing the translator and query router significantly reduces performance. For example, removing the query router leads to a severe decline in F1 score by 38.73\% and hit rate by 36.36\%. While removing post-processors results in a modest increase in precision by 2.62\%, it also increases execution time significantly, as depicted in Figure \ref{fig:exp4}.

\begin{table}[ht]
\centering
\caption{Performance Metrics of the Proposed Framework and Ablation Variants}
\label{tab:exp4}
\begin{tabular}{l>{\centering\arraybackslash}p{1.7cm}>{\centering\arraybackslash}p{1.7cm}>{\centering\arraybackslash}p{1.7cm}>{\centering\arraybackslash}p{1.7cm}}
\toprule
                                                  & \textbf{Precision} & \textbf{Recall} & \textbf{F1 score} & \textbf{Hit Rate} \\
\midrule
\textbf{Proposed Framework}                       & 80.56              & 58.46           & 66.21             & 93.64             \\
\midrule
\multirow{2}{*}{\textbf{Without Translator}}      & 54.20              & 45.98           & 47.22             & 92.73             \\
                                                  & $\downarrow$26.36             & $\downarrow$12.48          & $\downarrow$18.99            & $\downarrow$0.91             \\
\midrule
\multirow{2}{*}{\textbf{Without Router}}          & 55.91              & 19.56           & 27.48             & 57.27             \\
                                                  & $\downarrow$24.65             & $\downarrow$38.90          & $\downarrow$38.73            & $\downarrow$36.36            \\
\midrule
\multirow{2}{*}{\textbf{Without Post-Processors}} & 83.18              & 53.29           & 61.96             & 89.09             \\
                                                  & $\uparrow$2.62               & $\downarrow$5.17           & $\downarrow$4.25             & $\downarrow$4.55             \\
\bottomrule
\end{tabular}
\end{table}

Generally, the translator and query router are crucial for ensuring high accuracy, while post-processors help mitigate latency. Consequently, each component plays a distinct role in the framework's overall performance, and their absence leads to substantial performance degradation.

\section{Conclusions}

The proposed framework represents a significant advancement in archival search technologies by effectively leveraging the capabilities of LLMs to process input and output in natural language. LLMs are also successfully utilized to convert non-textual data into textual form, thereby augmenting existing metadata for richer semantic content and more efficient data storage. Our experiments validate the framework's effectiveness in improving search precision and relevance, particularly through the innovative use of multilingual LLMs and hybrid retrieval mechanisms. In our study, Mistral emerged as the most robust LLM, achieving a precision of 80.56\%. The ablation study underscores the critical roles of system components, revealing a precision drop of approximately 10-30\% without the translator/router and a time performance increase of around 20 seconds without post-processors on average. These findings indicate that each component contributes distinctly to overall performance and efficiency. Moving forward, this research opens pathways for further innovations in AI-driven archival systems, suggesting avenues for enhancement in multilingual search and system scalability. By continuously refining these technologies, we can significantly enhance the accessibility and utility of digital archives, meeting the evolving needs of users in the digital age.

\section*{Acknowledgment}
This research received support from the University of Information Technology, Vietnam National University of Ho Chi Minh City. We thank our colleagues from the Information System Laboratory who provided insight and expertise that greatly assisted the research.

%
%
%
\bibliographystyle{splncs04}
\bibliography{bibliography}

\end{document}